\documentclass[11pt,a4paper]{article}

\usepackage[]{times}
\usepackage[T1]{fontenc}
\usepackage[utf8]{inputenc}
\usepackage{textcomp} 
\usepackage{amsmath,amssymb}
\usepackage{csquotes}
\usepackage{booktabs}
\usepackage{datetime}
\usepackage{array}
\newcolumntype{P}[1]{>{\centering\arraybackslash}p{#1}}

\usepackage[english]{babel}

\usepackage[margin=1in]{geometry}
\usepackage{setspace}
\usepackage[hyphens]{url}
\usepackage{xcolor}
\usepackage[unicode=true]{hyperref}
\hypersetup{
  colorlinks=true,
  linkcolor={blue},
  pdfborder={0 0 0},
  colorlinks=true,
  linkcolor=.,
  citecolor=.,
  urlcolor=blue,
  breaklinks=true
}
\usepackage{listings}

\usepackage{graphicx}

\usepackage[style=apa,natbib]{biblatex}
\begin{document}

\begin{center}
 {\huge\bfseries Multivariate Time Series Classification: A Deep Learning Approach\\[5pt] } 
 \vspace{1.5cm}
 {\Large\bfseries Mohamed Abouelnaga, Julien Vitay, Aida Farahani}\\[5pt] 
 \vspace{1cm}

\end{center}

\section*{Abstract}

This paper investigates different methods and various neural network architectures applicable in the time series classification domain. The data is obtained from a fleet of gas sensors that measure and track quantities such as oxygen and sound. With the help of this data, we can detect events such as occupancy in a specific environment.
\\ At first, we analyze the time series data to understand the effect of different parameters, such as the sequence length, when training our models. These models employ Fully Convolutional Networks (FCN) and Long Short-Term Memory (LSTM) for supervised learning and Recurrent Autoencoders for semi-supervised learning. 
\\
Throughout this study, we spot the differences between these methods based on metrics such as precision and recall identifying which technique best suits this problem.

\newpage

\pagenumbering{roman} 
{
\hypersetup{linkcolor=}
\setcounter{tocdepth}{3}
\tableofcontents
}

\newpage

\pagenumbering{arabic} 

\section{Introduction}
\label{sec:introduction}
A time series is a collection of data points ordered in time \citep{1}. The analysis of this data is very beneficial in many domains, such as weather forecasting \citep{2}. An important application when we talk about time series classification is anomaly detection which is applicable in many domains, e.g., with the help of time series data such as velocity and acceleration, dangerous driving behaviors can be detected \citep{3}. \\

\subsection{Motivation}
\label{sec:Motivation}
Our motivation for this paper is to harness the time series data obtained from a fleet of gas sensors deployed by Corant GmbH / Air-Q company \footnote{\url{https://www.air-q.com}} in many homes and companies to detect events that can't be measured directly by these sensors. The primary function of these sensors is to measure and track many chemicals and quantities, such as O2, CO2, NO2, pressure, and sound. \\
With the help of machine learning, we extend these sensors' functionality by detecting more events, such as whether a specific environment is occupied within a particular time range. Also, we can see whether the windows of the place are open. We can notify the users of these events, which helps them to have more control over their environments and raises the safety level. Moreover, the investigated methods can be tailored to similar problems in the domain of time series analysis.

\subsection{Methods}
\label{sec:Methods}
Event detection in time series data can be done using various deep-learning architectures. We exploit the power of Fully Convolutional Networks (FCN) and Long Short-Term Memory (LSTM) in supervised learning. Also, we would introduce a simple Recurrent Autoencoder, which uses the unlabeled data in semi-supervised learning. \\
We mainly treat our problem as a multi-label classification in which we have two primary classes \{'person,' 'window\_open'\} that can be detected simultaneously, with binary cross entropy loss function \citep{5}. 
We also experiment with a separate network for each class in a single-label classification manner with softmax as an output layer \citep{6}.

\subsubsection{Fully Convolutional Network}
\label{sec:FCN}

\begin{figure}[htbp]
\centerline{\includegraphics[width=1\textwidth,height = 150 pt]{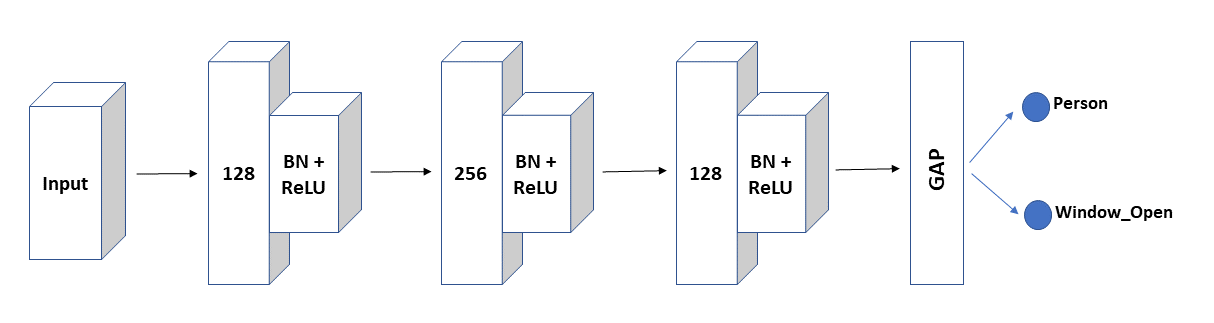}}
  \caption{Fully Convolutional Network (FCN).}
  \label{fig:FCN}
\end{figure}

Our problem deals with multivariate time series data, so FCN can be applied to grasp each input channel's local and global features. FCN has no pooling operations. Therefore it is used in other applications, such as semantic segmentation, to produce a pixel-wise output \citep{4}. \\
FCN performs as a feature extractor in our settings, as shown in Fig. \ref{fig:FCN}. FCN has several convolutional blocks, each consisting of a convolutional layer, followed by a batch normalization layer, and has Rectified Linear Unit (ReLU) as an activation function. The batch normalization helps to improve the overfitting and speed up the convergence \citep{8}.\\
As shown in \citep{7}, FCN originally stacks three convolutional blocks with 1-D kernel sizes of \{8, 5, 3\} and filters count of \{128, 256, 128\} respectively. The number of filters and kernel sizes can be optimized to better suit the problem, especially in small data sets.\\
Instead of applying directly a fully connected layer, the last convolutional block is followed by a Global Average Pooling (GAP) layer to average its output over time dimension. This enables a drastic reduction of the parameters.\\
To preserve the time series length after each convolutional block, the convolution operations have zero padding and a stride equal to 1.
One main advantage of FCN is that it can handle time sequences with different sizes, unlike the standard Recurrent Neural Networks that struggle with long-term dependencies.

\subsubsection{InceptionTime}
\label{sec:InceptionTime}

\begin{figure}[htbp]
\centerline{\includegraphics[width=0.9\textwidth]{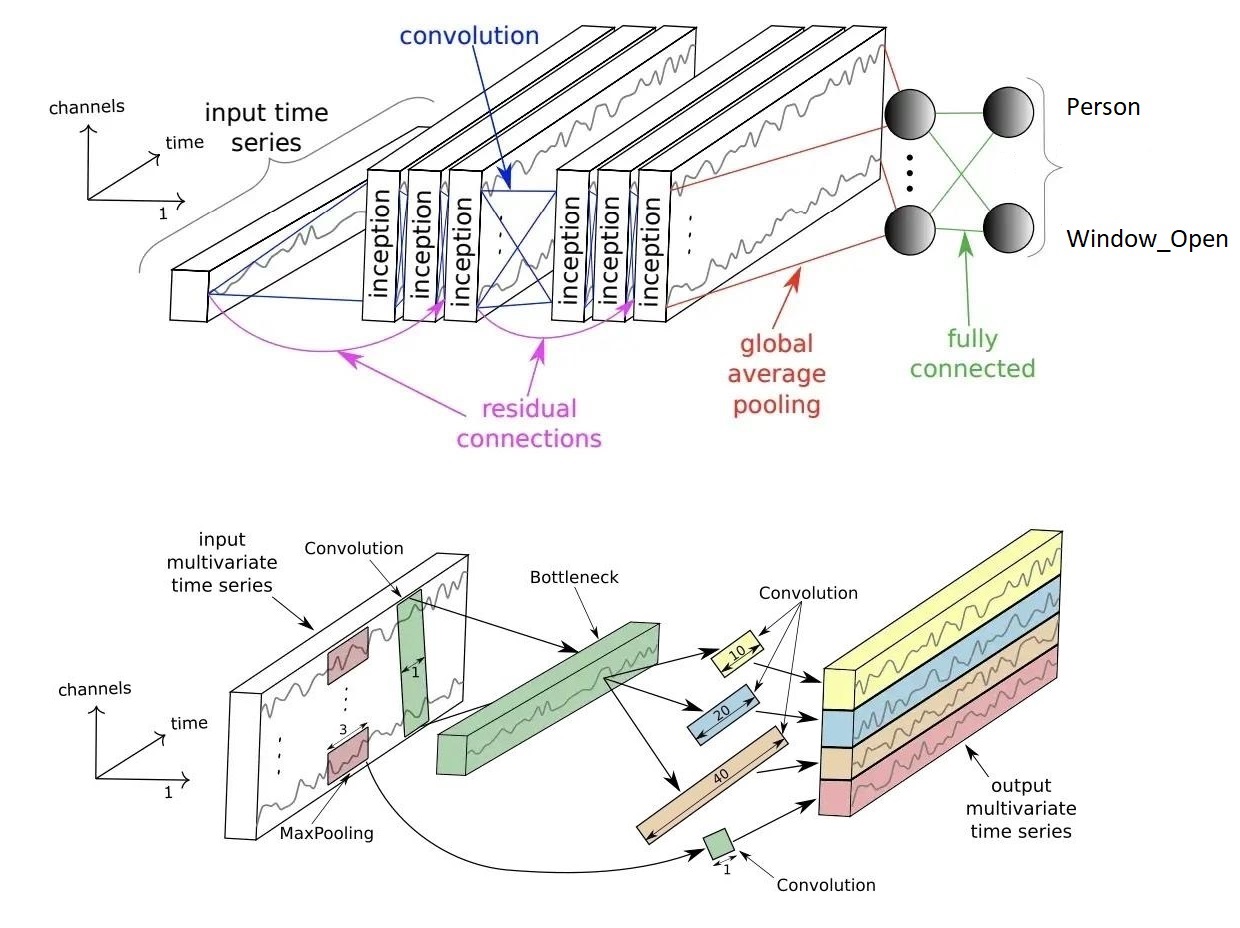}}
  \caption{Top: InceptionTime Network, Bottom: Single Inception Module. \\Source: \citep{9} }
  \label{fig:inceptiontime}
\end{figure}

InceptionTime \citep{9} is a state-of-art architecture that achieves very high accuracy when applied to time series classification. It is an ensemble of five Inception Networks that are initialized with different random weights with two residual blocks, as opposed to ResNet \citep{10}, which has three residual blocks as shown in Fig. \ref{fig:inceptiontime}. The residual connections fight the vanishing gradient problem \citep{11}.\\
Each residual block is comprised of three Inception modules. After the second block, a Global Average Pooling (GAP) is applied instead of directly using a fully connected layer.\\
As shown in Fig. \ref{fig:inceptiontime}, the core component of each Inception module is applying m filters with a stride equal to 1 and a length of 1. The result is called the bottleneck layer. This layer significantly reduces the dimension of the time series input and the model complexity. This technique allows for a longer filter with almost the same number of parameters as ResNet. \\
After that, several convolutions with different sizes are applied simultaneously on the bottleneck layer. To mitigate the perturbations, another MaxPooling layer is applied and concatenated with the output of the previous convolutions.

\subsubsection{Long Short-Term Memory}
\label{sec:lstm}

\begin{figure}[ht]
\centerline{\includegraphics[width=0.8\textwidth, height = 0.9\textwidth]{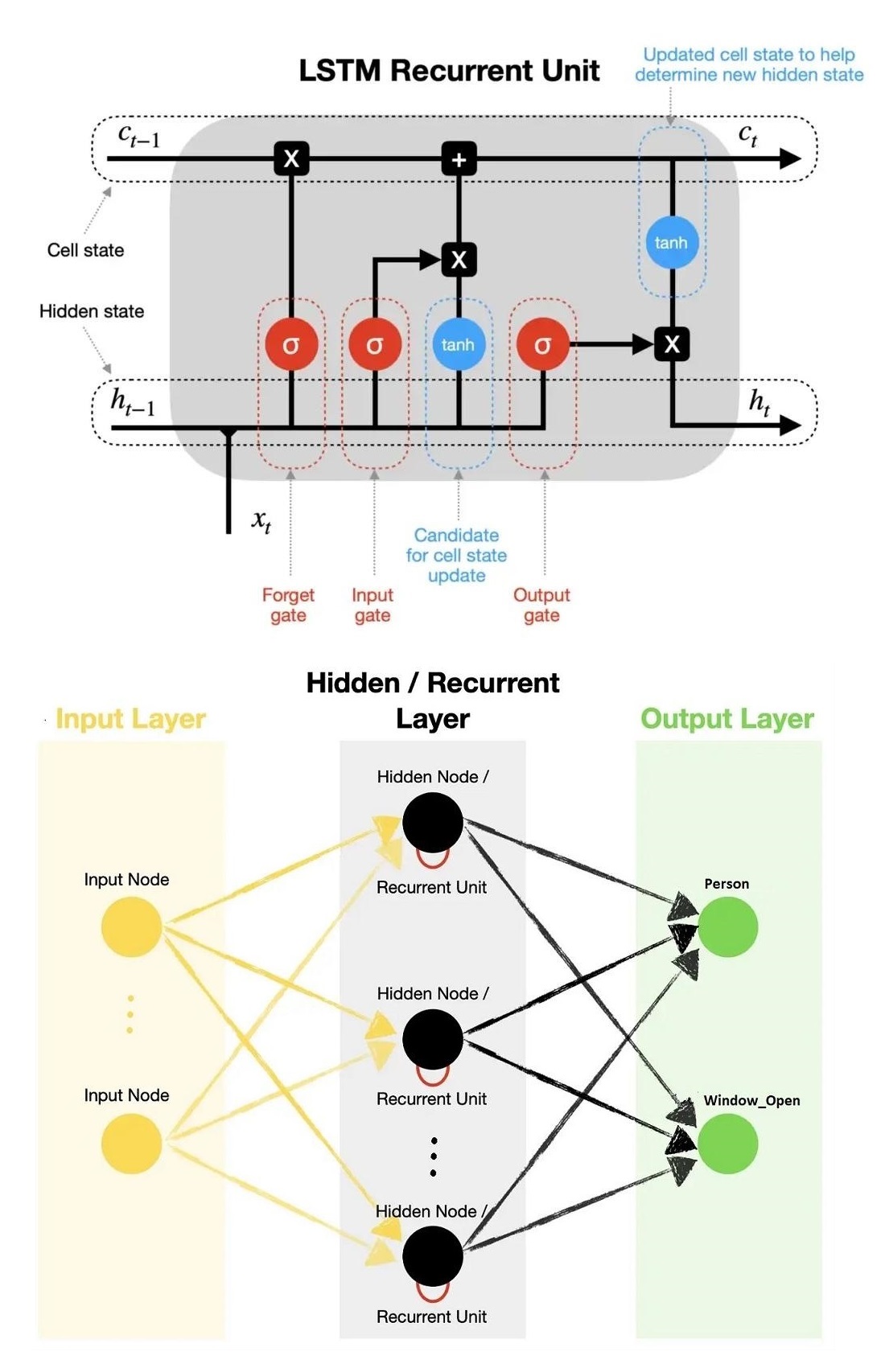}}
\caption{Top: LSTM Cell, Bottom: LSTM Network.\\Source: \url{https://towardsdatascience.com/lstm-recurrent-neural-networks-how-to-teach-a-network-to-remember-the-past-55e54c2ff22e} }
\label{fig:lstm}
\end{figure}

Recurrent Neural Network (RNN) is an essential architecture when dealing with time series data, as the output depends on a history of inputs ordered in time. However, RNN sufferers from detecting long-term dependencies due to the application of Back Propagation Through Time (BPTT) for a specific Horizon \citep{12}. \\
In contrast, Long Short-Term Memory (LSTM) \citep{13} cell uses a state which represents a "memory" or a "context" besides the inputs and the outputs to overcome this issue.  LSTM contains three gates to control the dependencies; an input gate to select the inputs, a forget gate to free some part of the memory, and an output gate to control the output, as shown in Fig. \ref{fig:lstm}. \\
We use an LSTM in our supervised method with only one hidden layer, as shown in Fig. \ref{fig:lstm}.

\subsubsection{Recurrent Autoencoder}
\label{sec:ae}

\begin{figure}[htbp]
\centerline{\includegraphics[width=1\textwidth]{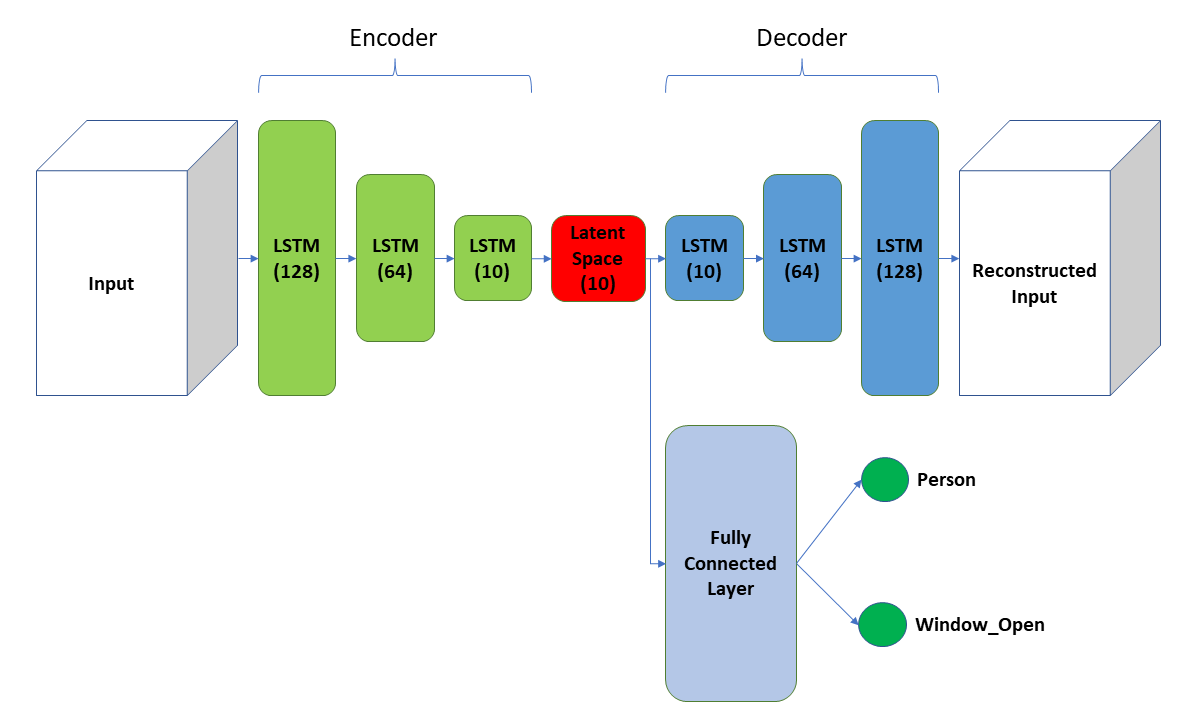}}
  \caption{Semi-supervised Learning using a Recurrent Autoencoder and a Shallow Classifier.}
  \label{fig:ae}
\end{figure}

Supervised learning algorithms require a lot of labeled data to train the model, especially if we have multivariate time series data. However, obtaining annotated data is a challenging and usually expensive task. According to Vapnik-Chervonenkis theorem \citep{14}, generalization error depends significantly on the amount of data the model is trained on, not only the complexity of the model.\\
As unlabeled data, in contrast, is cheap to obtain, we can combine them with a small amount of labeled data to get good accuracy in semi-supervised learning \citep{16}. Also, the random initialization of the parameters can lead to a longer training time. Therefore, as shown in Fig. \ref{fig:ae}, we can apply a Recurrent Autoencoder on the unlabeled data and minimize the reconstruction error, which is based on the Mean Squared Error (MSE) \citep{15}. \\
Then we use the Encoder only with frozen parameters with a shallow classifier on the labeled data, resulting in less number and good initialization of the parameters.

\subsection{Software Setup}
\label{sec:software}
For applying the previous neural network architectures, we used the "Tsai" library \citep{18}, which is based on "PyTorch" \citep{17} and "Fastai" \citep{19}.\\
For data manipulation and analysis, we used "Pandas" \citep{21}, "NumPy" \citep{20}, and "Scikit-Learn" \citep{22}.\\
For plotting the graphs, we used "Matplotlib" \citep{23} and "Plotly" \citep{24}.

\section{Experiments}
\label{sec:Experiments}
Before applying any method, we need to understand the data first. The data contains 17 features which are \{pressure, temperature, sound, tvoc, oxygen, humidity, humidity\_abs, co2, co, so2, no2, o3, pm2\_5, pm10, pm1, sound\_max, dewpt\} and two classes \{person, and window\_open\}. The sensors measure a sample every two minutes.\\
The labeled data is collected using only one device from \textit{July 2022} to \textit{December 2022}, while the unlabeled data is collected using 740 sensors over two years.

\subsection{Cleaning Data}
\label{sec:cleaning}
\begin{figure}[htbp]
\centerline{\includegraphics[width=1\textwidth]{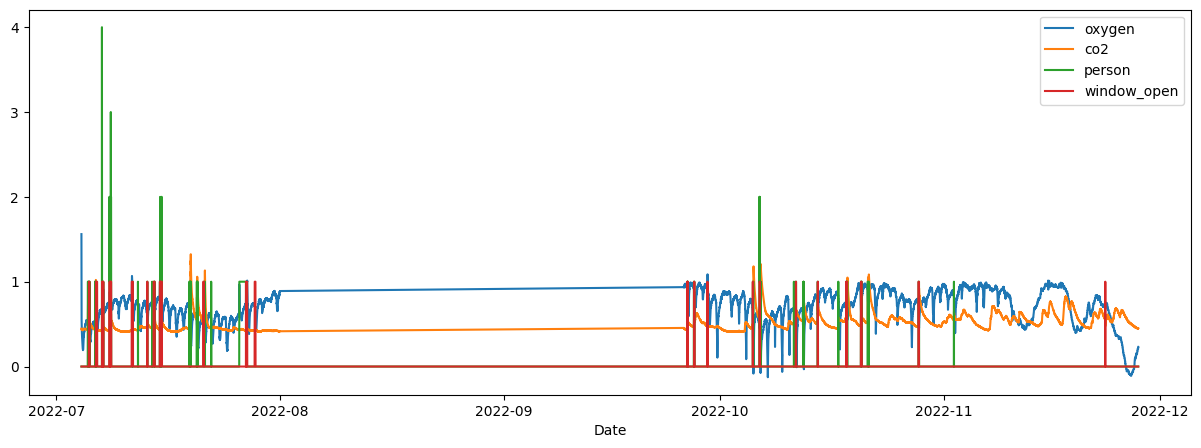}}
  \caption{Visualization of the labeled data.}
  \label{fig:Wholesignals}
\end{figure}

For a visualization of the labeled data (see Fig. \ref{fig:Wholesignals}), we used only two features \{o2,co2\} and the labels for the sake of readiness to spot how the classes are distributed over time. We note that no data was present in \textit{August} and most of \textit{September}.\\

\begin{figure}[htbp]
\centerline{\includegraphics[width=1\textwidth]{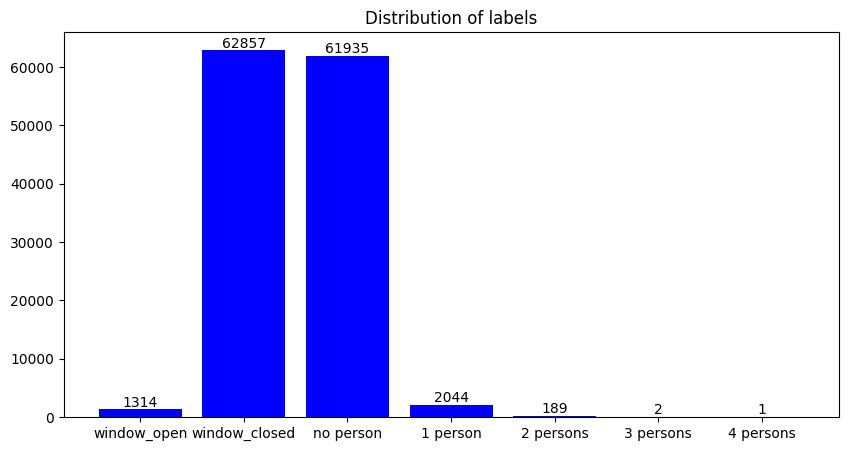}}
  \caption{Distribution of original labels in the labeled data.}
  \label{fig:original_labels}
\end{figure}

\begin{figure}[htbp]
\centerline{\includegraphics[width=0.8\textwidth]{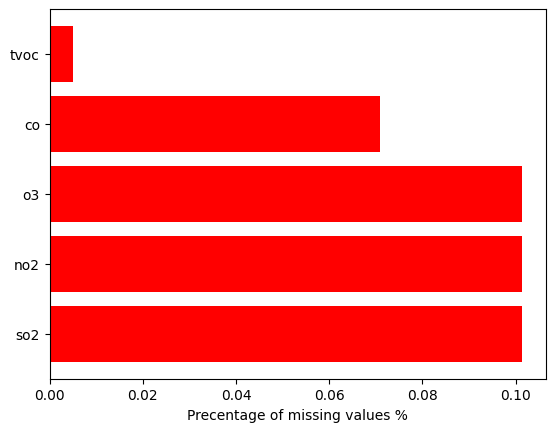}}
  \caption{Distribution of missing values in the labeled data.}
  \label{fig:nans}
\end{figure}

To better understand the distribution of labels, we can find in Fig. \ref{fig:original_labels} that the class person has a minimal number of labels when more than one person exists in the environment. Therefore, we merge all labels in which a person is found into one label. Hence we have two binary classes \{person, and Window\_open\}. \\
An important aspect when dealing with data is cleaning it from missing values; however, we shouldn't delete the missing values directly in time series data as that may affect the series frequency. \\ After we know the missing values in our data, as shown in Fig. \ref{fig:nans}, we can interpolate these values to keep the same timeline. Fortunately, a maximum of 20 missing values exist in our data, all in the beginning, so it was safe to delete them directly. However, we generally use linear interpolation to substitute for the missing values.

\subsection{Features Reduction}
\label{sec:Features}

\begin{figure}[htbp]
\centerline{\includegraphics[width=0.85\paperwidth]{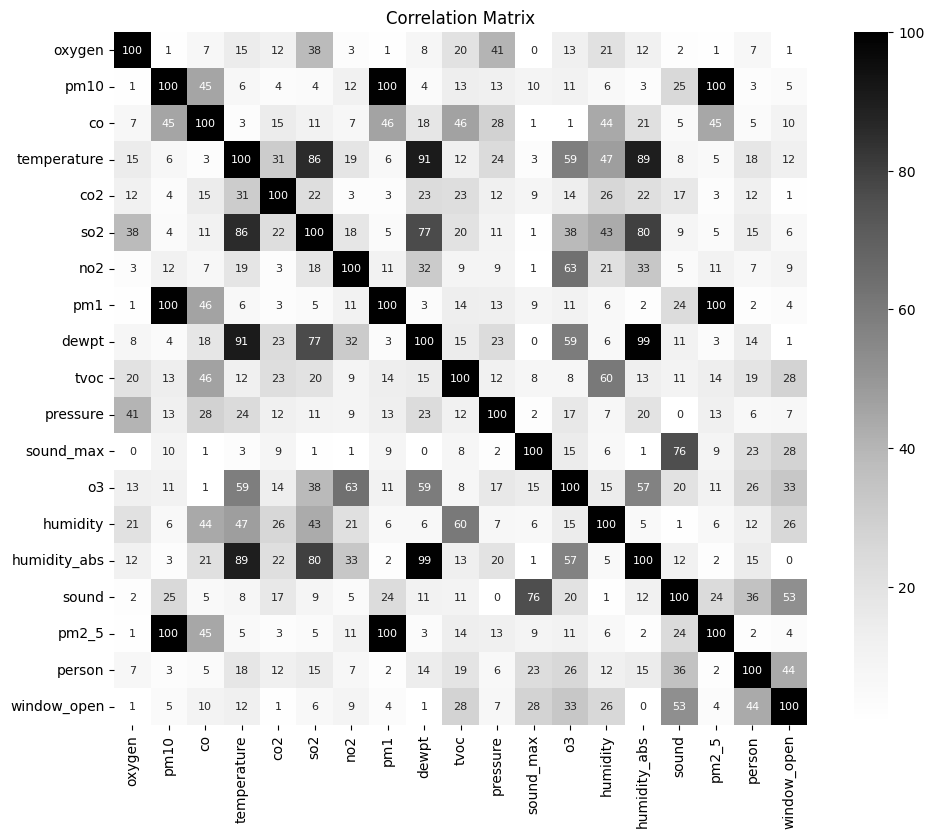}}
  \caption{Correlation Matrix of features and classes.}
  \label{fig:correlation}
\end{figure}

We have a small data set (below 70,000 samples), which can lead to low performance and high generalization errors. Also, not all features are equally crucial for accurate classification. One way to deal with that is to use fewer independent features.\\
We used Pearson correlation coefficient \citep{25} to build a correlation matrix as shown in Fig. \ref{fig:correlation} to eliminate the most correlated features. We obtained nine features \{humidity, temperature, tvoc, oxygen, co2, co, pressure, o3, sound\} which correlate well with the classes.

\subsection{Under Sampling}
\label{sec:sampling}

\begin{figure}[htbp]
\centerline{\includegraphics[width=1\textwidth]{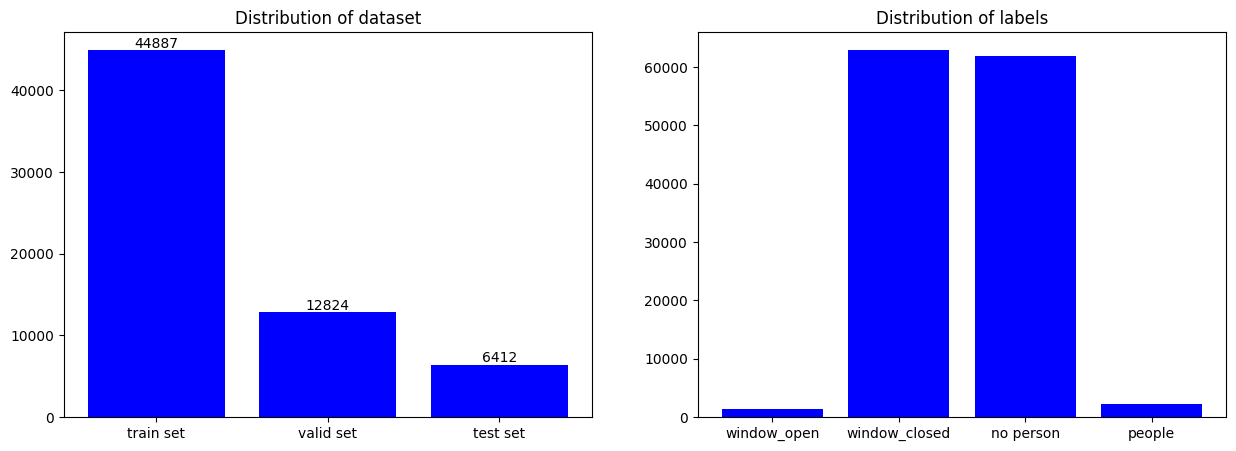}}
  \caption{Left: Distribution of unbalanced data set, Right: Distribution of unbalanced labels.}
  \label{fig:NoSamplingdata}
\end{figure}

After merging the labels of the data set, we can see an obvious imbalance of the labels as in Fig. \ref{fig:NoSamplingdata}. This can lead to false metrics when comparing different methods, e.g., for detecting a person; a model may produce high accuracy, although it performs poorly, as the ratio between the "People" label to "No Person" is very low.

\begin{figure}[t!]
\centerline{\includegraphics[width=1\textwidth]{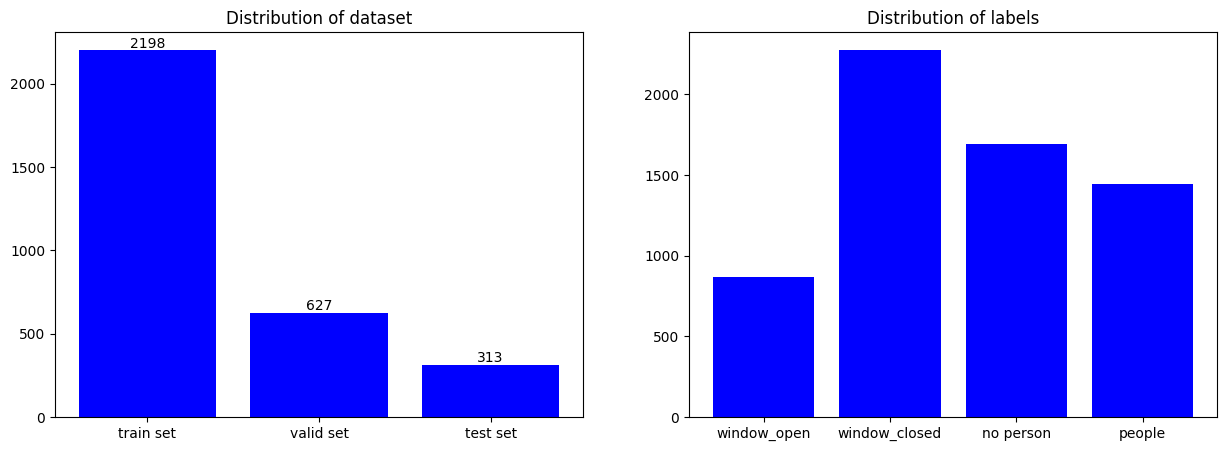}}
  \caption{Left: Distribution of data set, Right: Distribution of labels [After applying Under Sampling].}
  \label{fig:Samplingdata}
\end{figure}

We use under-sampling to overcome the imbalance problem and reflect more accurate metrics. As the count of a person's existence and open windows are very low, we choose a specific number of labels before and after every detection of a person or an open window. This leads to more balanced data, as shown in Fig. \ref{fig:Samplingdata}.

We note that we should apply the sliding operation on each segment resulting from under-sampling to obtain time sequences, then perform concatenation to maintain the time of the data.\\
To compare the results of under-sampling with the unbalanced data as shown in Table. \ref{tab:sampling}, we perform under-sampling with a size of 30 and train FCN with default parameters on 70\% of normalized data using standard scalar while keeping 20\% for validation and 10\% for testing as shown in Fig. \ref{fig:NoSamplingdata} and Fig. \ref{fig:Samplingdata}. This splitting is done randomly with a sequence length equal to 15, with a stride equal to one, and labeling at the end of each sequence.\\
We used only ten epochs with cosine learning rate scheduling \citep{26}. Also, we compared the usage of all features against the minimized features we deduced in the previous section.

\begin{table}[htbp]
\begin{center}

\begin{tabular}{P{3.5cm}|P{2cm}|P{2.2cm}|P{2cm}|P{1.6cm}|P{1.7cm}}

& \textbf{\textit{No. of parameters }}& \textbf{\textit{Training time (s)}} & \textbf{\textit{Test accuracy }} & \textbf{\textit{F1 score (person)}}  & \textbf{\textit{F1 score (window)}} \\
 \hline
\textbf{\textit{All features, no sampling}}  & 275,970 & 60 & 99\% & 0.93 & 0.94 \\
 \hline
 \textbf{\textit{All features, sampling}}  & 275,970 & 11 & 90\% & 0.90 & 0.94 \\
\hline
\textbf{\textit{Minimal features, no sampling}}  & 269,698 & 63 & 98\% & 0.90 & 0.93 \\
\hline
 \textbf{\textit{Minimal features, sampling}}  & 269,698 & 11 & 90\% & 0.91 & 0.93 \\
\hline
\end{tabular}
\end{center}
\caption{Performance of under-sampling and minimized features}
\label{tab:sampling}
\end{table}

We can see from the results that using unbalanced data for training consumes ($ \approx $ 6x) training time compared to under-sampling, with similar F1 scores. Hence, we can use under-sampling to save time. Also, we can safely choose a minimal number of features.

\subsection{Sequence Labeling and Normalization}
\label{sec:Normalization}

We segment samples into sequences of a specific length for time series data and label each sequence. This label can be the first label, the very end, or the mean value of labels of all samples in each sequence. \\
In Table. \ref{tab:segments}, we compare the performance of FCN with the same settings as before while using different sequence labeling. We can find that labeling the sequences from the start or taking the mean value would result in better performance.

\begin{table}[htbp]
\begin{center}
\begin{tabular}{P{2.2cm}|P{2cm}|P{2cm}|P{2cm}|P{1.6cm}|P{1.7cm}}
& \textbf{\textit{Train loss }}& \textbf{\textit{Valid loss}} & \textbf{\textit{Test accuracy }} & \textbf{\textit{F1 score (person)}}  & \textbf{\textit{F1 score (window)}} \\
 \hline
\textbf{\textit{First label}}  & 0.1 & 0.1 & 94\% & 0.94 & 0.96 \\
 \hline
\textbf{\textit{ Mean label}} & 0.07 & 0.06 & 96\% & 0.96 & 0.97 \\
\hline
\textbf{\textit{Last label}}  & 0.14 & 0.12 & 90\% & 0.91 & 0.93 \\
\hline
\end{tabular}
\end{center}
\caption{Performance of different Sequence Labeling}
\label{tab:segments}
\end{table}

Also, Table. \ref{tab:normalization} compares applying a standard scalar and min-max scalar \citep{27} when normalizing the data with the mean value for sequence labeling. We obtain similar results. Hence we will use a standard scalar from now on. 

\begin{table}[htbp]
\begin{center}
\begin{tabular}{P{3cm}|P{2cm}|P{2cm}|P{2cm}|P{1.6cm}|P{1.7cm}}
& \textbf{\textit{Train loss }}& \textbf{\textit{Valid loss}} & \textbf{\textit{Test accuracy }} & \textbf{\textit{F1 score (person)}}  & \textbf{\textit{F1 score (window)}} \\
 \hline
\textbf{\textit{Standard scalar}} & 0.08 & 0.06 & 95\% & 0.96 & 0.97 \\
 \hline
\textbf{\textit{Min-Max scalar}}  & 0.07 & 0.05 & 96\% & 0.96 & 0.97 \\
\hline
\end{tabular}
\end{center}
\caption{Standard Scalar vs. Min-Max Scalar}
\label{tab:normalization}
\end{table}

\subsection{Benchmarking Architectures}
\label{sec:Benchmarking}

As shown in Table. \ref{tab:benchmarking}, we compared FCN, LSTM (uni-directional and bi-directional one-layer Networks), and InceptionTime, all with default parameters with the mean value for sequence labeling. Regardless of the good results of InceptionTime, we would skip using it later as we have a small amount of labeled data compared to the parameter count of the model. Also, it has longer training time. From now on, we will use FCN and a uni-directional one-layer LSTM as the main models.

\begin{table}[htbp]
\begin{center}
\begin{tabular}{P{3cm}|P{2cm}|P{2.2cm}|P{2cm}|P{2cm}|P{2cm}}
& \textbf{\textit{No. of parameters }} & \textbf{\textit{Training time (s)}} & \textbf{\textit{Valid Accuracy }}& \textbf{\textit{Train loss}} & \textbf{\textit{Valid loss }} \\
 \hline
\textbf{\textit{InceptionTime}} & 456,258 & 19 & 98\% & 0.05 & 0.04 \\
 \hline
\textbf{\textit{FCN}}  & 269,698 & 9 & 97\% & 0.08 & 0.07 \\
\hline
\textbf{\textit{Uni-directional LSTM}}  & 43,802 & 8 & 97\% & 0.07 & 0.09 \\
\hline
\textbf{\textit{Bi-directional LSTM}}  & 87,602 & 9 & 97\% & 0.09 & 0.1 \\
\hline
\end{tabular}
\end{center}
\caption{Benchmarking FCN, LSTM, and InceptionTime}
\label{tab:benchmarking}
\end{table}

\subsection{Minimized Architecture and Sequence Length}
\label{sec:length}

\begin{figure}[htbp]
\centerline{\includegraphics[width=1\textwidth]{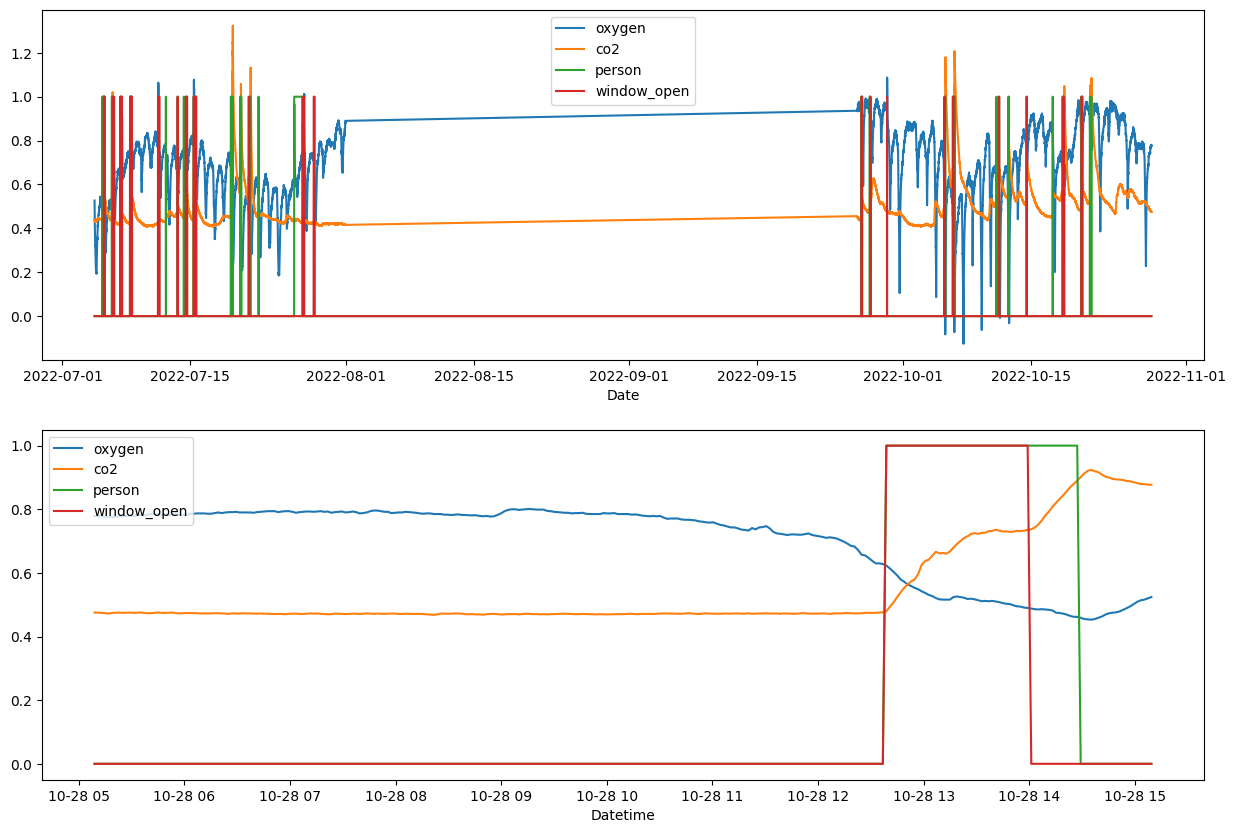}}
  \caption{Top: A separate training set, Bottom: A separate test set.}
  \label{fig:traintestNovember}
\end{figure}

FCN with default parameters has a very high parameter count (269,698) compared to training samples, even after increasing the under-sampling size to 50 (5,000 training sequences on average, depending on sequence labeling). Therefore, we initially tune FCN to consist of only two convolutional blocks with kernel sizes of \{5, 3\} and filter count of \{16, 32\}, resulting in 2,418 parameters. This can mitigate overfitting.\\
Also, the test set was chosen using random splitting after segmentation, which may lead to data leakage while training, producing more biased accuracy. Therefore, we prefer a different test set that is separated before segmentation.\\
Fig. \ref{fig:traintestNovember} shows this test set and the remaining training set. Using this test set, we experiment with various sequence lengths using FCN and 100 epochs for training with early stopping \citep{29}. According to the results shown in Table. \ref{tab:lengths}, we can choose a sequence length of 7 as it represents 14 minutes in real-time to avoid a long prediction time; also, we choose the f st label for each sequence to perform further experiments. \\

Also, we would keep considering our problem as a multi-label classification as it is more realistic as a person and an open window can be detected simultaneously; also, the results are comparable with single-label classification, e.g., if we take each sequence with a length equals to 10 with a mean value labeling, we will obtain F1 scores of (0.94, 0.84) for person and window classes respectively.

\begin{table}[htbp]
\begin{center}
\begin{tabular}{P{2.2cm}|P{4cm}|P{3.5cm}|P{3.5cm}}
& \textbf{\textit{Sequence length = 15 }} & \textbf{\textit{F1 score (person)}}  & \textbf{\textit{F1 score (window)}} \\
 \hline
\textbf{\textit{First label}}  &  &  0.94 & 0.90 \\
 \hline
 \textbf{\textit{First label}}  &  &  0.96 & 0.92 \\
 \hline
 \hline
 & \textbf{\textit{Sequence length = 10 }} & \textbf{\textit{F1 score (person)}}  & \textbf{\textit{F1 score (window)}} \\
 \hline
\textbf{\textit{First label}}  &  &  0.92 & 0.84 \\
 \hline
 \textbf{\textit{First label}}  &  &  0.95 & 0.85 \\
 \hline
 \hline
  & \textbf{\textit{Sequence length = 7 }} & \textbf{\textit{F1 score (person)}}  & \textbf{\textit{F1 score (window)}} \\
 \hline
\textbf{\textit{First label}}  &  &  0.96 & 0.90 \\
 \hline
 \textbf{\textit{First label}}  &  &  0.96 & 0.92 \\
 \hline
\end{tabular}
\end{center}
\caption{Comparing different sequence lengths}
\label{tab:lengths}
\end{table}

\subsection{Hyperparameter Optimization}
\label{sec:optimization}

We use Optuna \citep{28}, an automatic hyperparameter optimization framework to optimize hyperparameters. The search space for the minimized FCN would be the number of filters from 8 to 32 with a step of 4. And for LSTM, we optimize the hidden size from 10 to 30 with a step of 2, also the dropout from 0.1 to 0.5 with a step of 0.1. We maximize the F1 score for optimization using 100 Optuna trials. \\ 
That leads to an FCN of 2,306 parameters with two convolutional blocks, with {32,8} filters of sizes {5,3}, respectively. Moreover, we get a one-layer uni-directional LSTM of 2,950 parameters with a hidden size of 26 and a dropout of 0.2. \\
We present the results in Table. \ref{tab:optuna}, with precision and recall metrics \citep{30} beside the F1 score to reflect the contribution of false positives and false negatives separately.\\

\begin{table}[htbp]
\begin{center}
\begin{tabular}{P{2cm}|P{5cm}|P{5cm}}
& \textbf{\textit{Precision, Recall, F1 score (person) }}& \textbf{\textit{Precision, Recall, F1 score (person)}} \\
 \hline
\textbf{\textit{FCN}}  & 0.91, 1.0, 0.95 & 1.0, 0.97, 0.98 \\
 \hline
 \textbf{\textit{LSTM}}  & 0.90, 1.0, 0.94 & 0.82, 1.0, 0.90 \\
 \hline
\end{tabular}
\end{center}
\caption{FCN vs. LSTM after hyperparameter optimization}
\label{tab:optuna}
\end{table}

\begin{figure}[htbp]
\centerline{\includegraphics[width=0.9\textwidth]{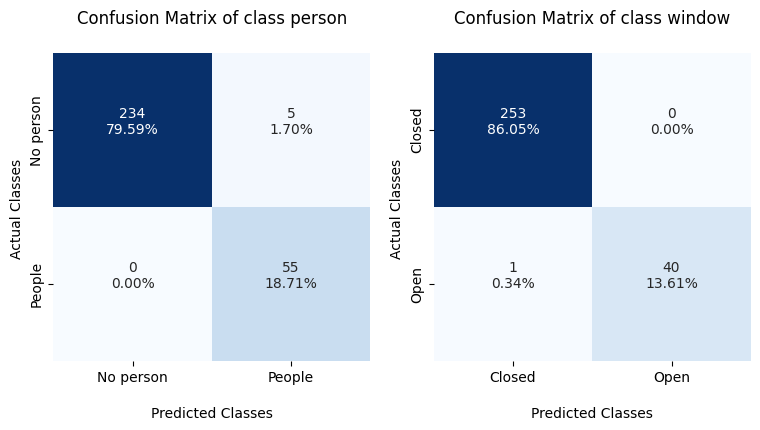}}
  \caption{Confusion matrices of FCN.}
  \label{fig:confusion_FCN}
\end{figure}

\begin{figure}[htbp]
\centerline{\includegraphics[width=0.9\textwidth]{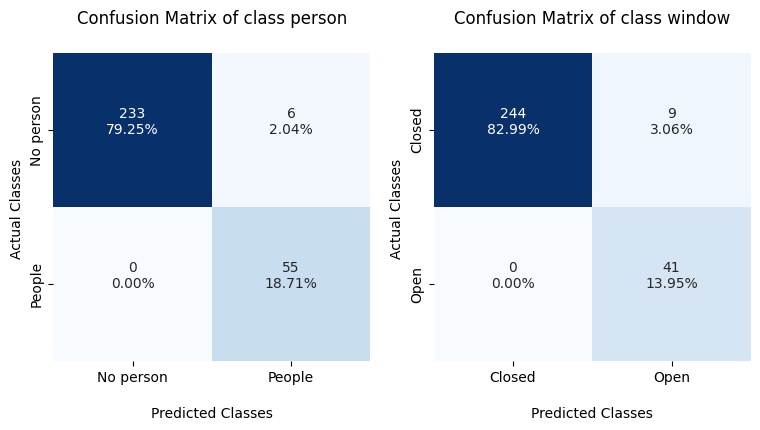}}
  \caption{Confusion matrices of LSTM.}
  \label{fig:confusion_LSTM}
\end{figure}

For better records of true positives, true negatives, false positives, and false negatives, Fig. \ref{fig:confusion_FCN} and Fig. \ref{fig:confusion_LSTM} show the confusion matrices in case of FCN and LSTM, respectively, for both classes \{person, window\_open\}.

\subsection{Predictions Distribution and Features Visualization}
\label{sec:Distribution}

\begin{figure}[htbp]
\centerline{\includegraphics[width=0.9\textwidth]{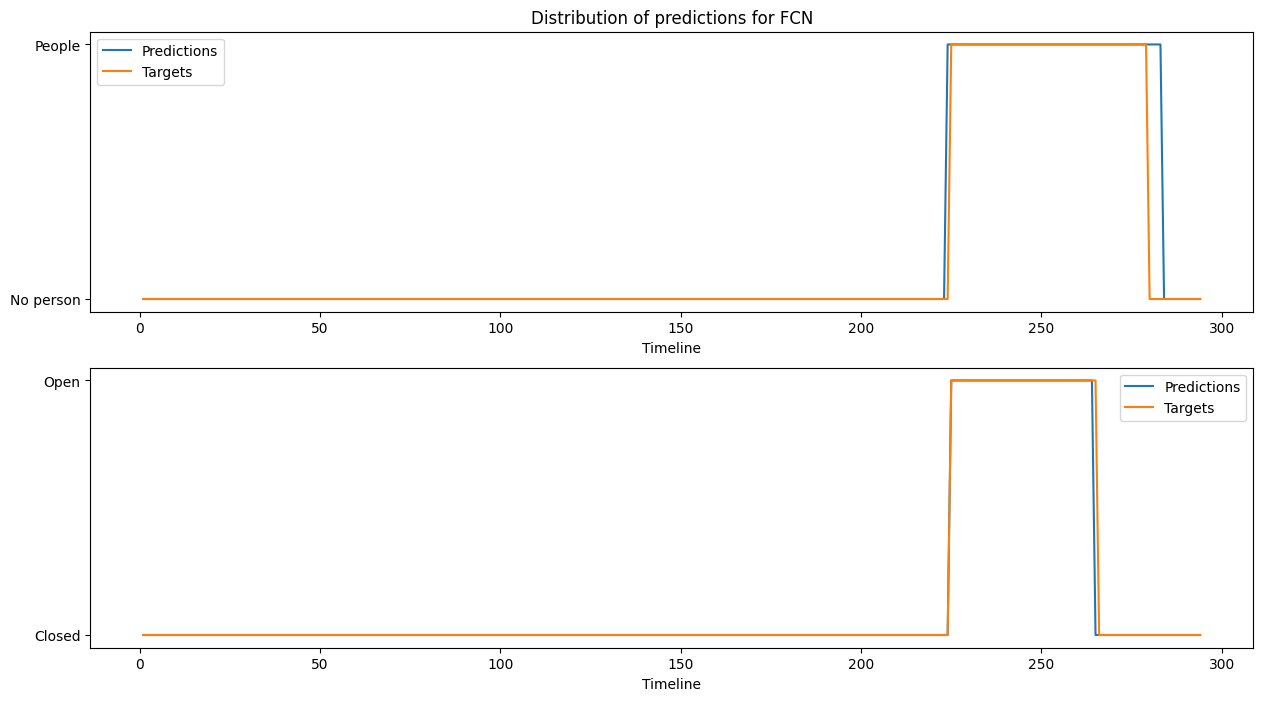}}
  \caption{Distribution of predictions for FCN.}
  \label{fig:predictionsFCN}
\end{figure}

\begin{figure}[htbp]
\centerline{\includegraphics[width=0.9\textwidth]{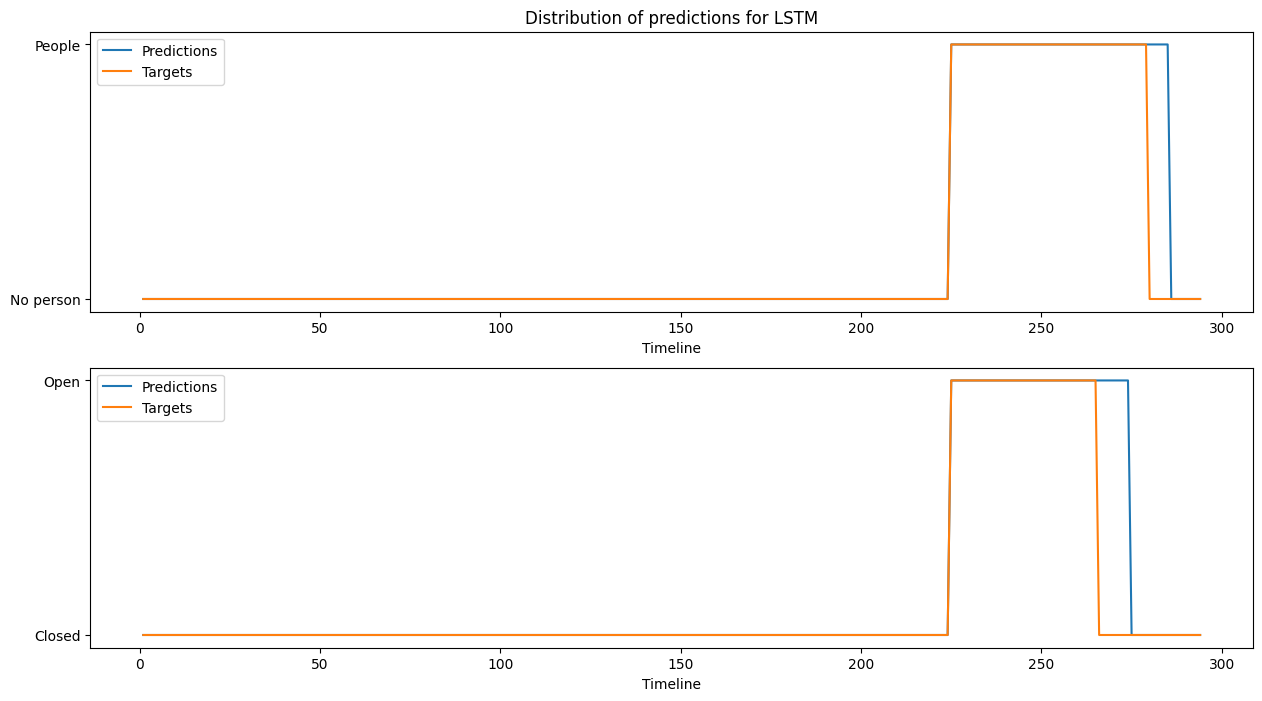}}
  \caption{Distribution of predictions for LSTM.}
  \label{fig:predictionsLSTM}
\end{figure}

\begin{figure}[htbp]
\centerline{\includegraphics[width=0.9\textwidth]{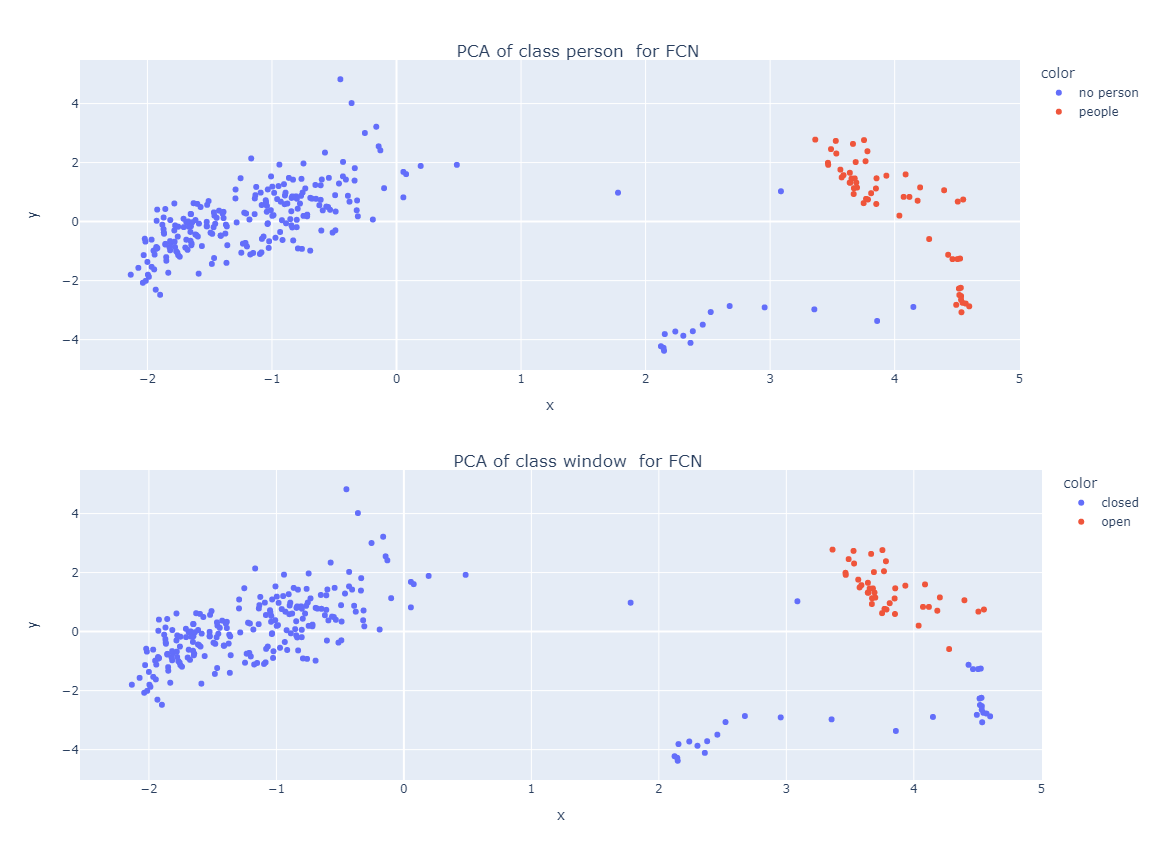}}
  \caption{PCA for FCN with labeled data.}
  \label{fig:pcaFCN}
\end{figure}

\begin{figure}[htbp]
\centerline{\includegraphics[width=0.9\textwidth]{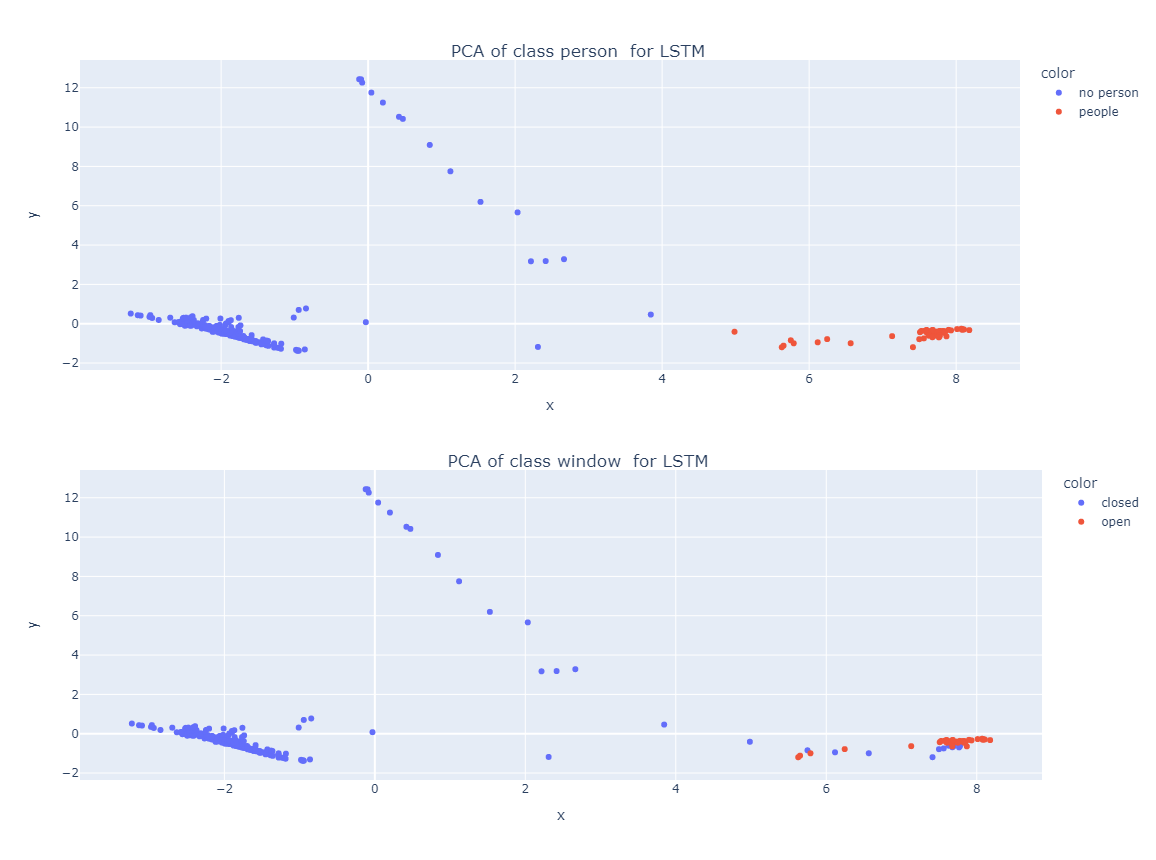}}
  \caption{PCA for LSTM with labeled data.}
  \label{fig:pcaLSTM}
\end{figure}

\begin{figure}[htbp]
\centerline{\includegraphics[width=0.9\textwidth]{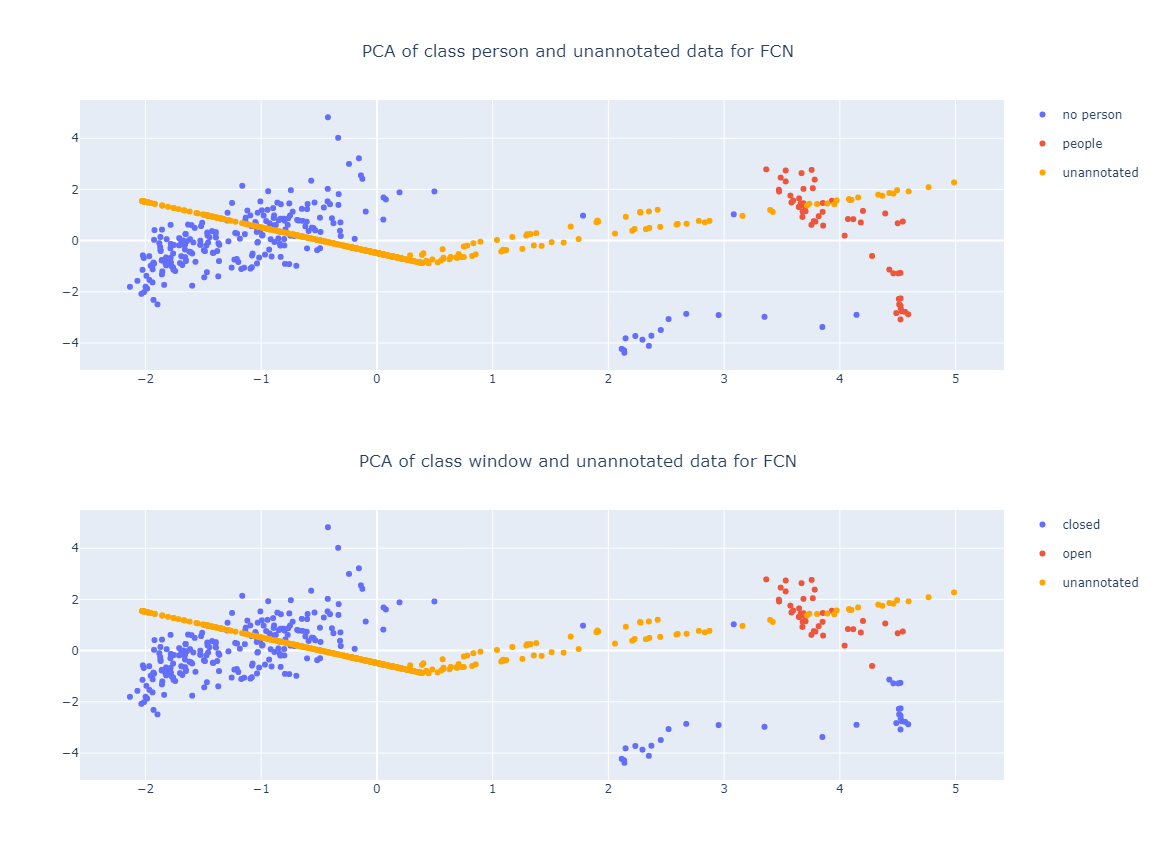}}
  \caption{PCA for FCN with labeled and unlabeled data.}
  \label{fig:pcaFCNUnannotated}
\end{figure}

As precision and recall don't reflect the distribution of predictions over time, we visualize this distribution in Fig. \ref{fig:predictionsFCN} and Fig. \ref{fig:predictionsLSTM} for FCN and LSTM, respectively.
Also, we visualize the feature space using Principle Component Analysis (PCA) \citep{31} for FCN and LSTM as in \ref{fig:pcaFCN} and Fig. \ref{fig:pcaLSTM}, respectively, on our separate labeled test set. \\
We can also see, as in Fig. \ref{fig:pcaFCNUnannotated} when using unlabeled data from a different sensor, it does not follow the same distribution as in the labeled test set.

\subsection{Encoder Classifier}
\label{sec:Encoder}

\begin{figure}[htbp]
\centerline{\includegraphics[width=0.9\textwidth]{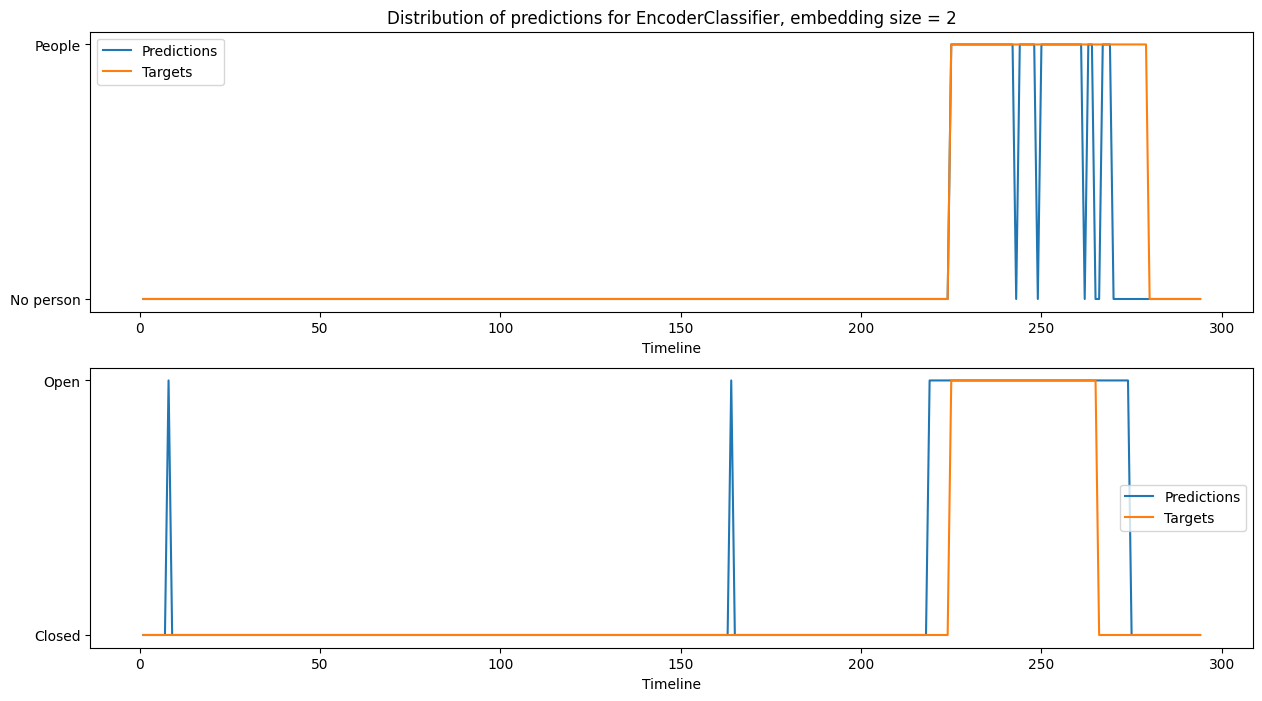}}
  \caption{Distribution of predictions for encoder classifier with latent\_size = 2.}
  \label{fig:predictionsEncoder2}
\end{figure}

\begin{figure}[htbp]
\centerline{\includegraphics[width=0.9\textwidth]{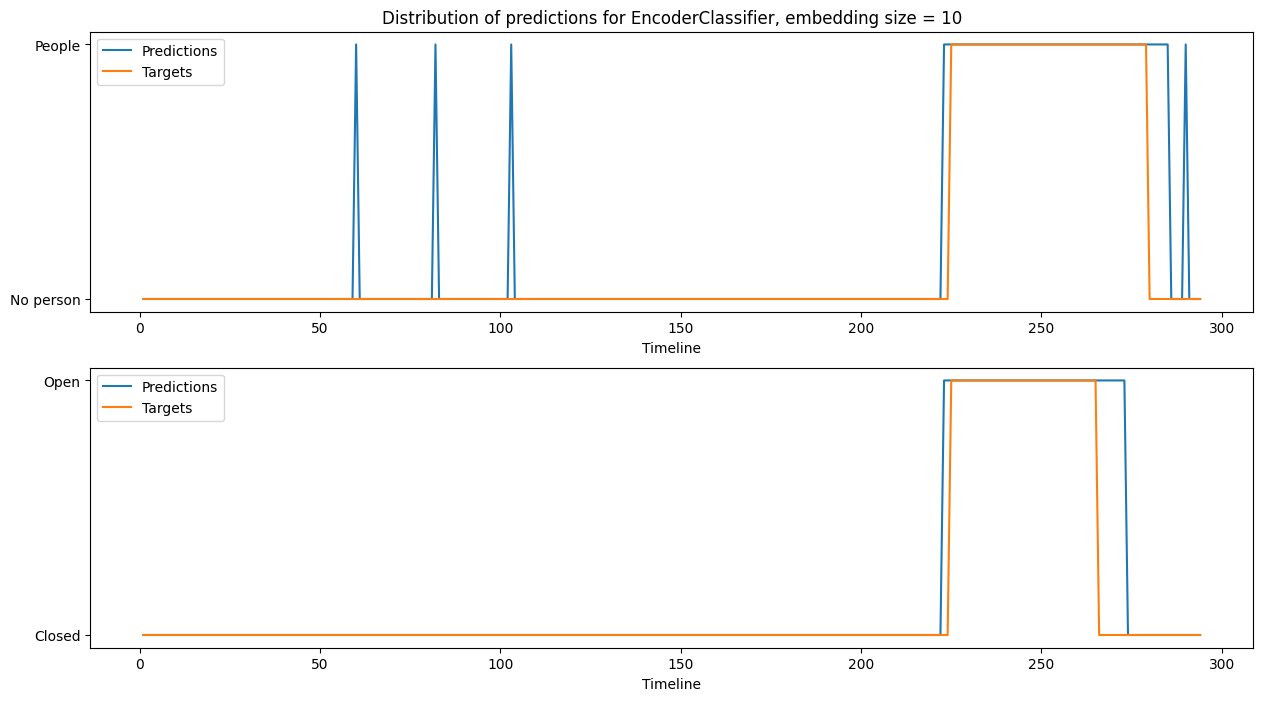}}
  \caption{Distribution of predictions for encoder classifier with latent\_size = 10.}
  \label{fig:predictionsEncoder10}
\end{figure}

\begin{figure}[htbp]
\centerline{\includegraphics[width=0.9\textwidth]{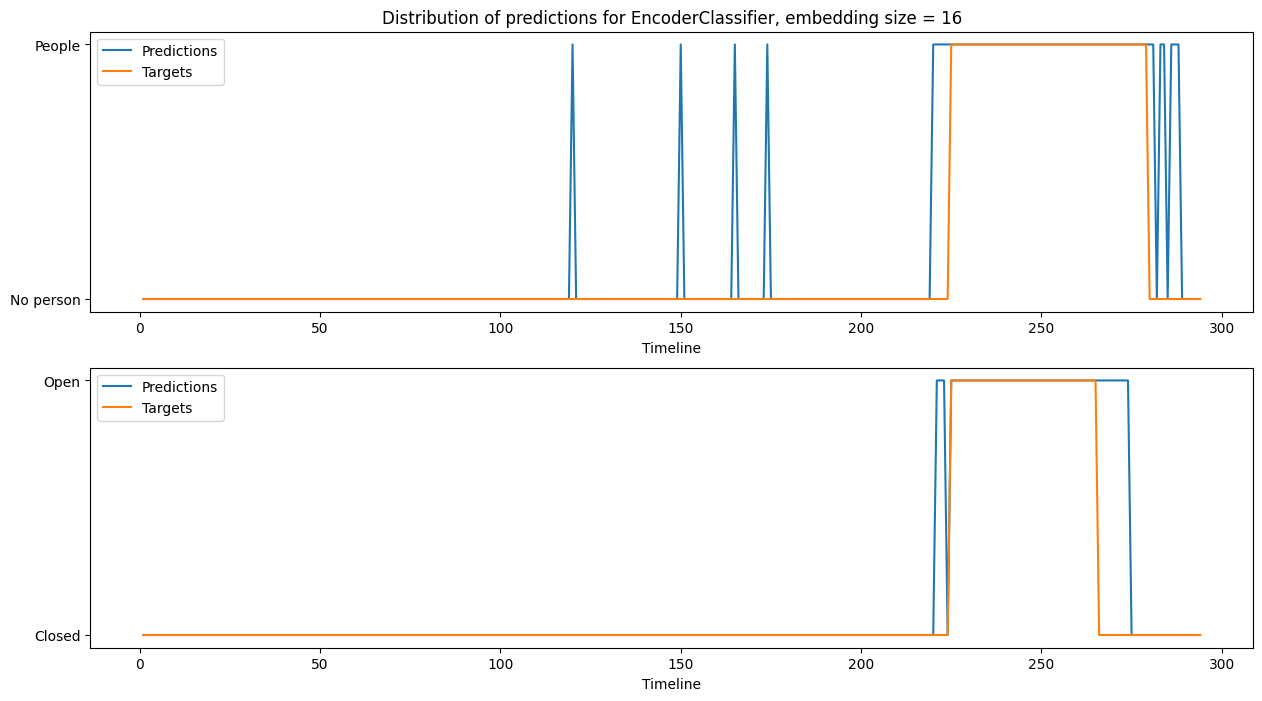}}
  \caption{Distribution of predictions for encoder classifier with latent\_size = 16.}
  \label{fig:predictionsEncoder16}
\end{figure}

Now we can use around 8,000,000 unlabeled sequences to train the recurrent autoencoder, then use the trained encoder with a shallow classifier consisting of a fully connected layer of 100 neurons and train it on the labeled data. \\
The encoder consists of three LSTM layers with sizes of \{128, 64, latent\_size\} respectively, and the decoder also consists of three LSTM layers with sizes of \{latent\_size, 64, 128\} respectively; where "latent\_size" represents the latent space size.
We experiment with three different latent space sizes of \{2, 10, 16\} with parameters of \{244,449, 249,825, 254,865\} respectively.\\
After testing the different encoder classifiers on the labeled test set we used before, we can see in Table. \ref{tab:encoderclassifer} that shows the scores of the encoder classifier, also in Fig. \ref{fig:predictionsEncoder2}, Fig. \ref{fig:predictionsEncoder10}, and Fig. \ref{fig:predictionsEncoder16} which show the distribution of predictions; that the embedding size of two is shallow to compress the 17 features efficiently. \\
Therefore we will conduct the next experiments using the embedding size of 10 as it has similar results with 16 but with fewer parameters.\\

\begin{table}[htbp]
\begin{center}
\begin{tabular}{P{2cm}|P{5cm}|P{5cm}}
\textbf{\textit{Latent\_size}} & \textbf{\textit{Precision, Recall, F1 score (person) }}& \textbf{\textit{Precision, Recall, F1 score (person)}} \\
 \hline
\textbf{\textit{2}}   & 1.0, 0.72, 0.83 & 0.70, 1.0, 0.82 \\
 \hline
 \textbf{\textit{10}}   & 0.82, 1.0, 0.90 & 0.80, 1.0, 0.89 \\
 \hline
  \textbf{\textit{16}}   & 0.77, 1.0, 0.87 & 0.77, 1.0, 0.87 \\
 \hline
\end{tabular}
\end{center}
\caption{Performance of encoder classifiers with various latent space sizes}
\label{tab:encoderclassifer}
\end{table}

\begin{figure}[htbp]
\centerline{\includegraphics[width=0.9\textwidth]{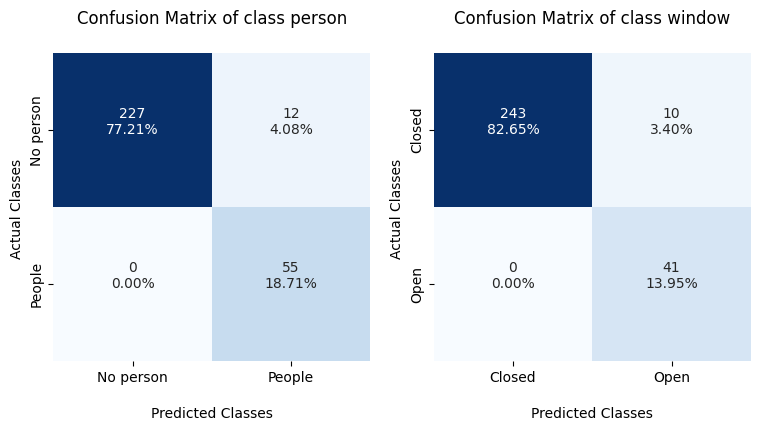}}
  \caption{Confusion matrices of encoder classifier with latent\_size = 10.}
  \label{fig:confusion_AE10}
\end{figure}

\begin{figure}[b!]
\centerline{\includegraphics[width=0.9\textwidth]{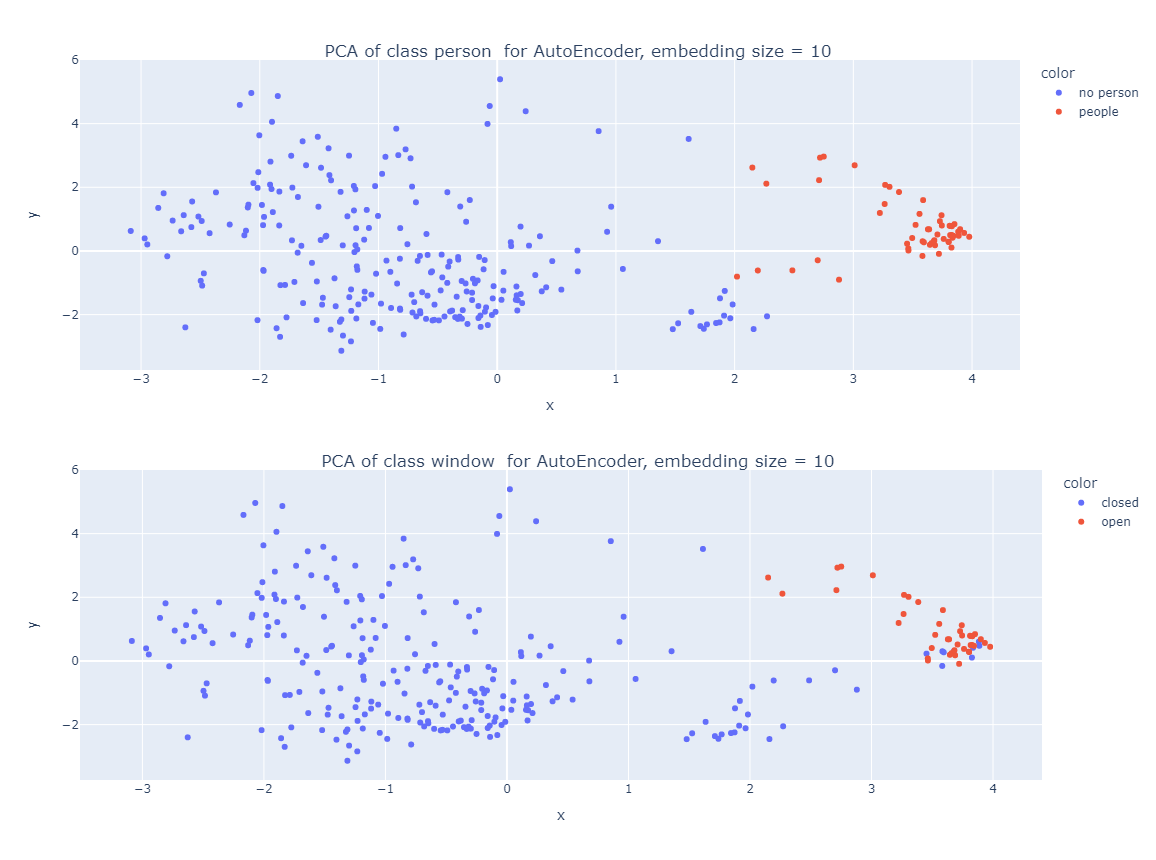}}
  \caption{PCA for encoder classifier with latent\_size = 10 with labeled data.}
  \label{fig:pcaAutoEncoder}
\end{figure}

We also show in Fig. \ref{fig:confusion_AE10} the confusion matrices for the encoder classifier with latent\_size = 10, also Fig. \ref{fig:pcaAutoEncoder} shows the PCA of the latent space of the Encoder.

\begin{figure}[htbp]
\centerline{\includegraphics[width=0.9\textwidth]{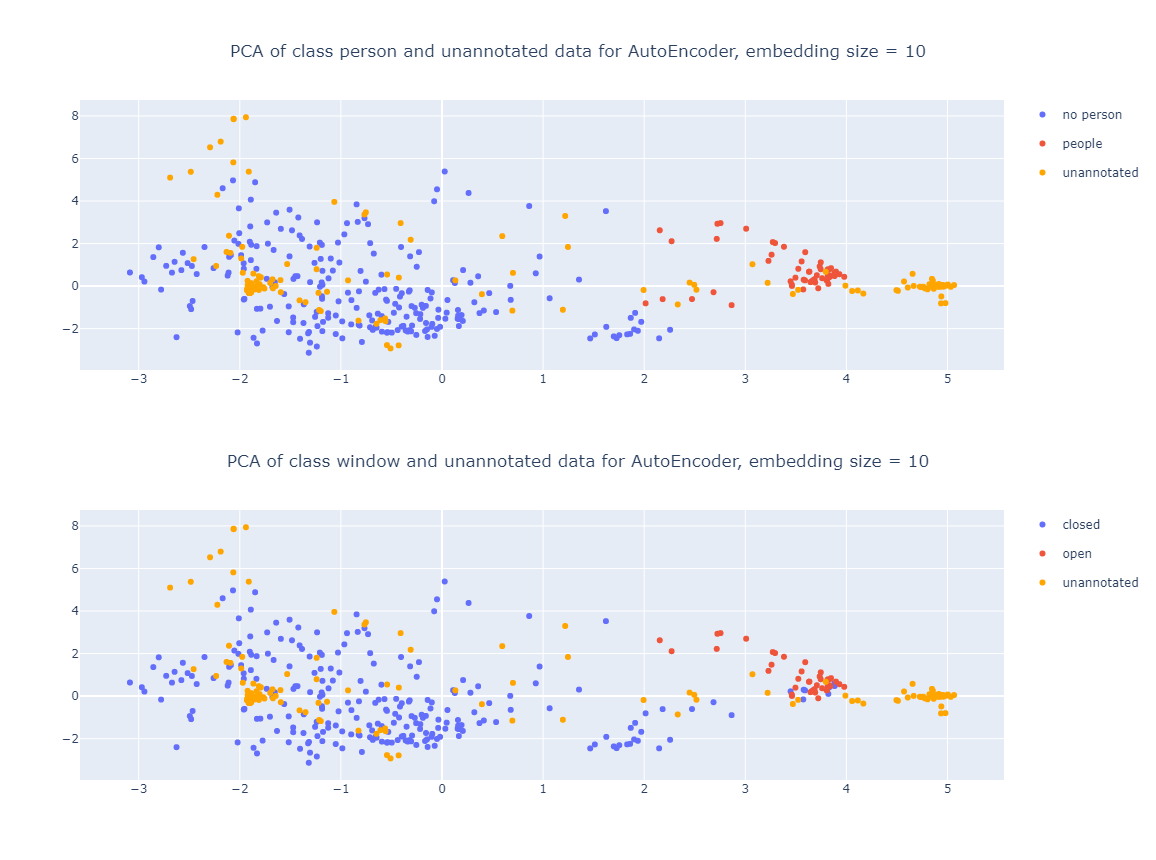}}
  \caption{PCA for for encoder classifier with latent\_size = 10 with labeled and unlabeled data.}
  \label{fig:pcaAutoEncoderunlabeled}
\end{figure}

Also, we can see in Fig. \ref{fig:pcaAutoEncoderunlabeled}, which shows the PCA of the encoder classifier when applied to the same unlabeled test set we used before in the PCA of FCN, that it follows the distribution of the feature space in contrast to FCN.

\begin{figure}[b!]
\centerline{\includegraphics[width=0.9\textwidth]{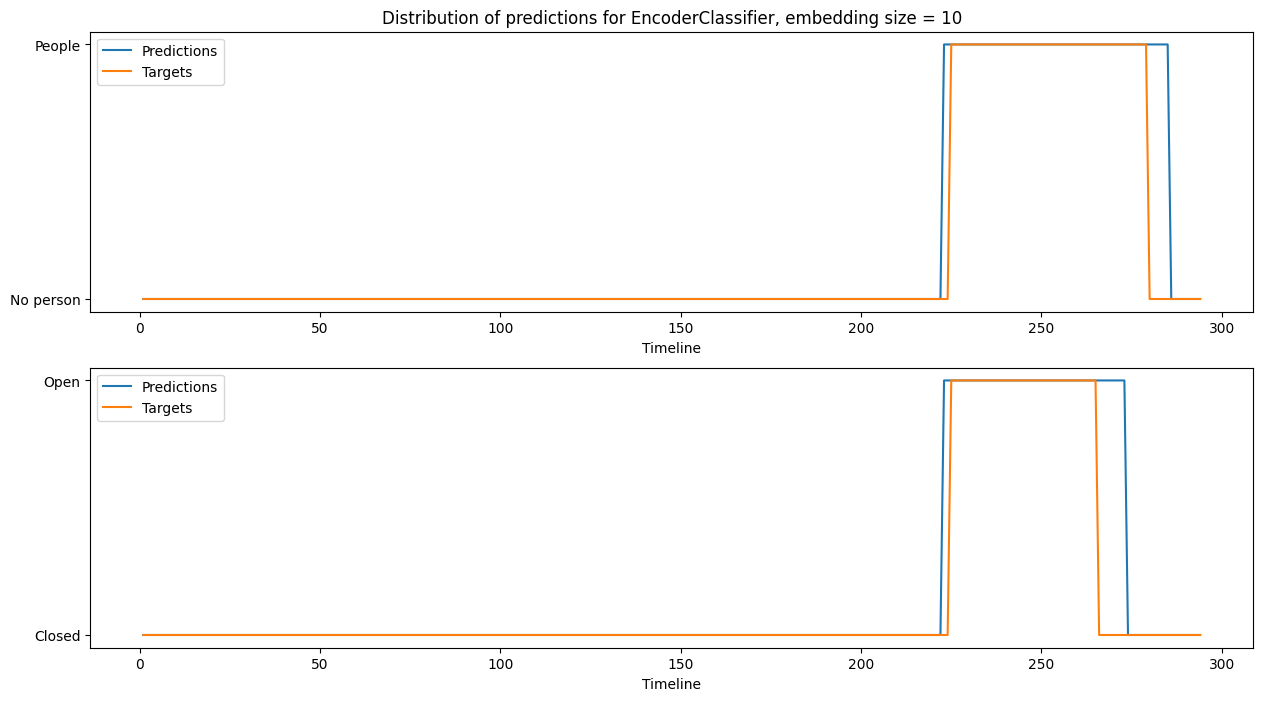}}
  \caption{Smoothed distribution of predictions for encoder classifier with latent\_size = 10.}
  \label{fig:predictionsEncoder10Smoothed}
\end{figure}

We can also smooth the predictions by rectifying the spikes with different widths that are considered errors, as shown in Fig. \ref{fig:predictionsEncoder10Smoothed}. \\

\begin{figure}[htbp]
\centerline{\includegraphics[width=1\textwidth]{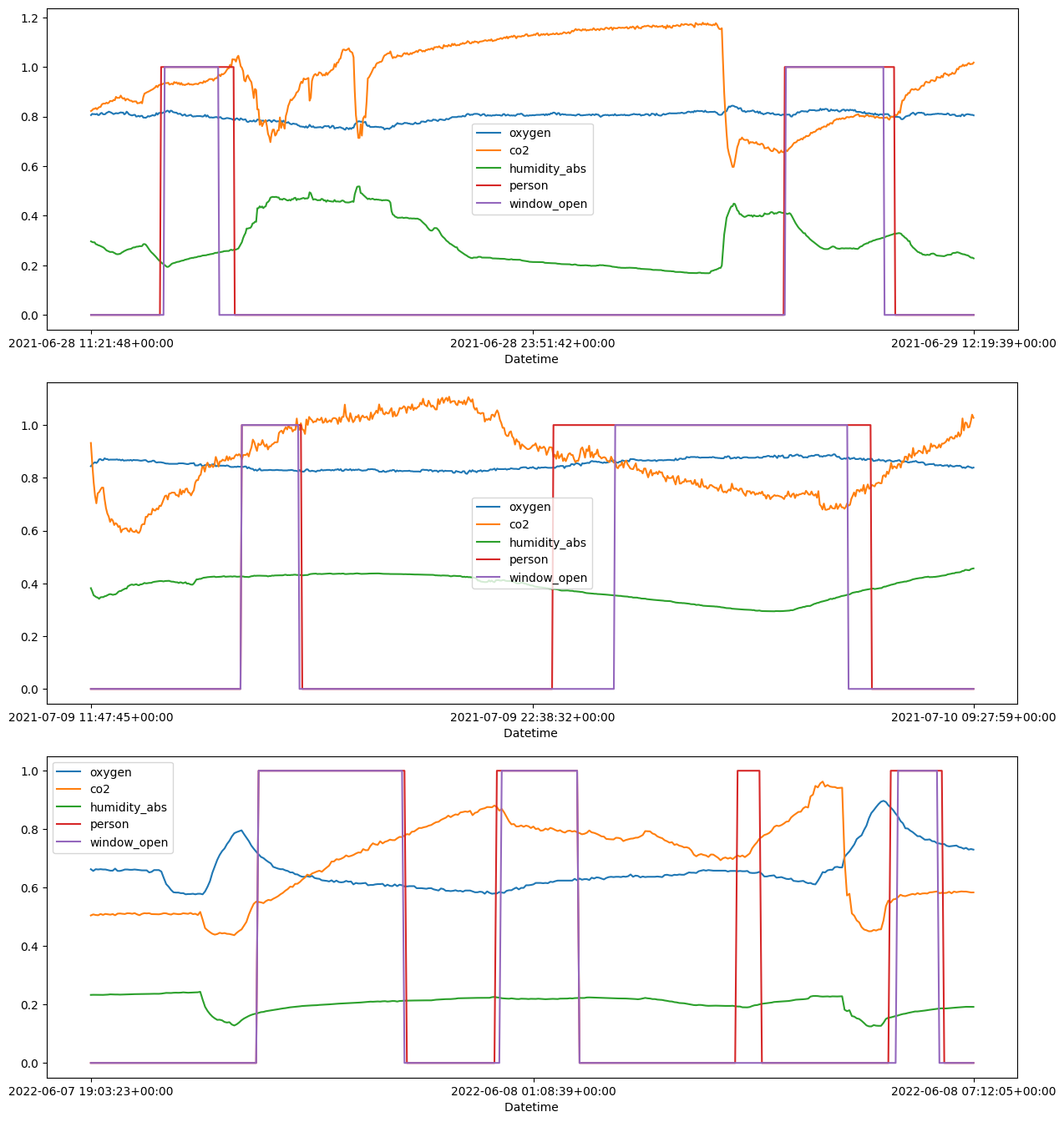}}
  \caption{Smoothed distribution of predictions and some signals when applying encoder classifier with latent\_size = 10 on unlabeled test sets from different sensors.}
  \label{fig:signals_AE}
\end{figure}

At last, we include some smoothed predictions as shown in Fig. \ref{fig:signals_AE} using the encoder classifier when applied to various unlabeled test sets collected from different sensors combined with only three signals \{o2, co2, humidity\_abs\} for better visualization. The signals are included to show the correlation with the predictions over time.

\section{Conclusion}
\label{sec:Conclusion}

Time series data can be found anywhere in nature; therefore, using them benefits many applications. In our paper, we presented two different deep-learning approaches to notify a user if a person exists or the window is open in his environment using data obtained from various gas sensors. In the first approach, we used supervised learning using two architectures which are FCN and LSTM. This method works well but needs more generalization if data is not sufficient.\\
Also, we examined the usage of a semi-supervised learning technique by training a recurrent autoencoder on the unlabeled data, then using the trained encoder with a shallow classifier on the labeled data. This allows using less labeled data as we train only the classifier while freezing the encoder.\\
We should take care of some practices before dealing with time series data, such as cleaning data by interpolating missing values and not directly removing them to reserve the timeline. Also, choosing a sequence length and the labeling position for each sequence are two important factors. There is no significant difference between standard and min-max scalars as long as the normalization step is performed before training. Also, analyzing the feature space for the used architecture and visualizing the distribution of predictions give more insights into the best-fitting solution.\\
Ultimately, we can get more robust results if we use more data, in-depth hyperparameter optimization, or even different architectures. One important future architecture to examine is the self-supervised learning technique using Transformers \citep{32}.

\newpage
\printbibliography

\end{document}